\documentclass[preprint,review,12pt]{elsarticle}

\usepackage{amssymb}
\usepackage{graphicx}
\usepackage{amsmath}
\usepackage{amssymb}
\usepackage{booktabs}
\usepackage{multirow}
\usepackage{bm}
\usepackage{rotating}

\journal{Pattern Recognition}

\begin{document}

\begin{frontmatter}


 \author[1]{Hyeono Jung}
 \ead{jho42555@sogang.ac.kr}

 \author[1]{Jangwon Lee}
 \ead{ljangw9719@sogang.ac.kr}

\author[1]{Jiwon Yoo}
 \ead{z1u0131@sogang.ac.kr}

  \author[1]{Dami Ko}
 \ead{dibidimi@sogang.ac.kr}
 
 \author[1]{Gyeonghwan Kim\corref{cor1}}
 \ead{gkim@sogang.ac.kr}
 \ead[url]{https://mmi.sogang.ac.kr/mmi/}

 \cortext[cor1]{Corresponding author}

 \affiliation[1] {organization={Dept. of Electronic Engineering},
             addressline={Sogang University},
             city={35 Baekbum-ro, Mapo-gu, Seoul 04107},
             country={Rep. of Korea}}

\title{PAFormer : Part Aware Transformer for Person Re-identification}




\begin{abstract}

Within the domain of person re-identification (ReID), partial ReID methods are considered mainstream, aiming to measure feature distances through comparisons of body parts between samples. However, in practice, previous methods often lack sufficient awareness of anatomical aspect of body parts, resulting in the failure to capture features of the same body parts across different samples. To address this issue, we introduce \textbf{Part Aware Transformer (PAFormer)}, a pose estimation based ReID model which can perform precise part-to-part comparison. In order to inject part awareness to pose tokens, we introduce learnable parameters called `pose token' which estimate the correlation between each body part and partial regions of the image. Notably, at inference phase, PAFormer operates without additional modules related to body part localization, which is commonly used in previous ReID methodologies leveraging pose estimation models. Additionally, leveraging the enhanced awareness of body parts, PAFormer suggests the use of a learning-based visibility predictor to estimate the degree of occlusion for each body part. Also, we introduce a teacher forcing technique using ground truth visibility scores which enables PAFormer to be trained only with visible parts. A set of extensive experiments show that our method outperforms existing approaches on well-known ReID benchmark datasets.
\end{abstract}


\begin{highlights}
\item Previous partial ReID methods' problem; not performing part-to-part comparison
\item Introduce pose estimation based ReID model 
\item Localization module free in inference phase
\item Alleviate an occlusion problem by learned visibility predictor
\end{highlights}


\end{frontmatter}


\section{Introduction}
\label{Introduction}


Person re-identification (ReID) is a type of image retrieval task that aims to find samples with the same ID as a given query image from a gallery set. Specifically, the methodology termed part-based ReID or partial ReID aims to achieve identification by comparing distinctive features extracted from specific body parts. Following the success of stripe-based methods~\cite{PCB,MGN,LATransformer}, which compute feature distances between spatially aligned regions after dividing images according to a predetermined division rule, the emergence of partial ReID methodologies has commenced in earnest.

Recently, methods using attention mechanisms for localizing body parts have been proposed~\cite{ABDNet,Deeply,HarmoAttn}, with ViT~\cite{ViT}-based methods~\cite{Nformer,TransReID,HAT} being encompassed in this category. These approaches typically employ part prototypes~\cite{PAT,AAformer,PFD,DCFormer} which are expected to represent specific body parts, to capture partial features. In these methods, they use loss term which combines cross-entropy loss and triplet loss. However, this loss term is more inclined towards guiding the prototype to excel in ID prediction within the training set rather than learning awareness for the human anatomy. Consequently, there is a significant risk of overfitting, potentially leading the model to capture only dominant information from individual training samples.

\begin{figure}[t!]
    \begin{center}
    \includegraphics[width=\columnwidth]{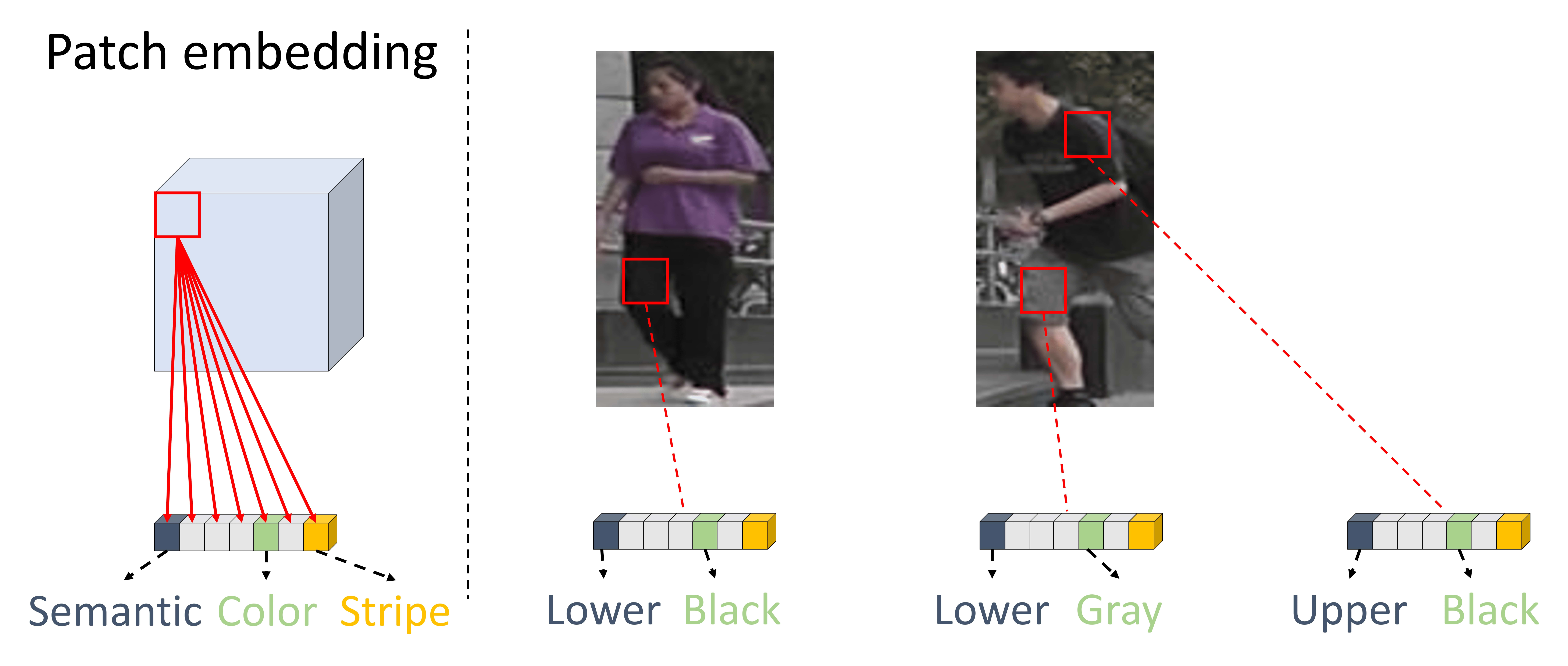}
    \end{center}
    \caption{\textbf{Left)} Patch embedding in ViT : CNN filters of the same size as the patch are applied, extracting visual representation features. \textbf{Right)} The generated patch embeddings contain both body part-related channels and unrelated channels.}
    \label{Fig1} 
\end{figure}

We elaborate the aforementioned issues frequently observed in attention-based ReID with its variant, ViT-based model. In ViT-based models, each part prototype gathers information based on the cosine similarity with patch tokens, allowing all channels of the token to contribute to the result. Therefore, in case there are channels in the token unrelated to human body, the resulting similarity map may not accurately represent body parts. As in Figure~\ref{Fig1}, if there exists a channel related with color, the black lower body part of the first sample might show a higher similarity with the black upper body part of the second sample, rather than the gray lower body part of that. 

\begin{figure}[t!]
    \begin{center}
    \includegraphics[width=\columnwidth]{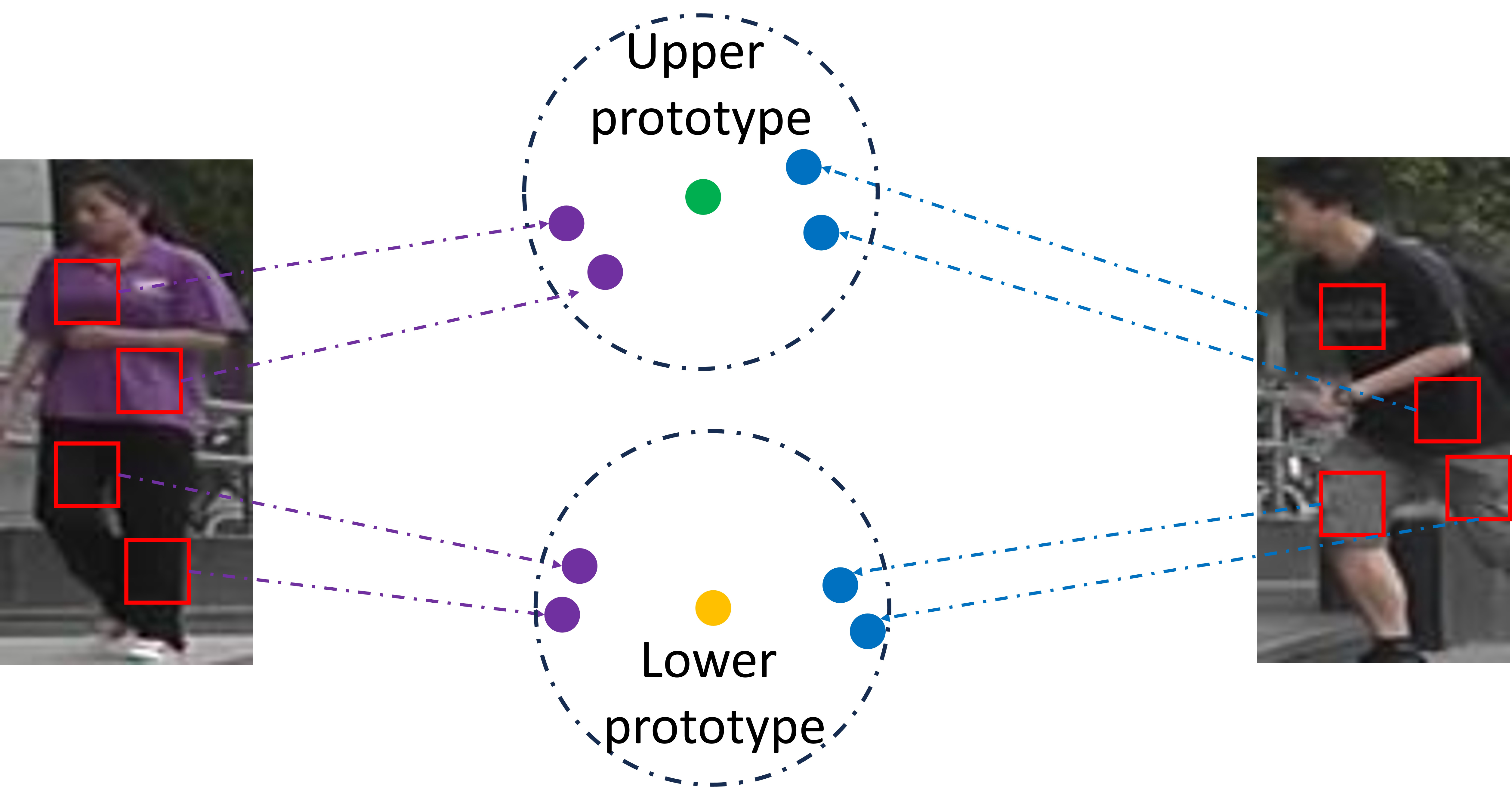}
    \end{center}
    \caption{Each part tokens should be projected close to the patch tokens corresponding to the body part it represents in the embedding space. If body part-unrelated features are involved in the similarity computation of ViT, there is a risk of projecting based on specific appearances rather than body part.}
    \label{Fig2} 
\end{figure}

In partial ReID, we expect the projection of prototypes and patch tokens in the embedding space to occur as depicted in Figure~\ref{Fig2}. Many methods leveraging human body information~\cite{PFD,ISP,BPBReid} have been proposed to achieve this behavior. However, these methods often entail challenges, such as the reliance on external pose estimation models even during inference phase, or the introduction of additional body part localization modules beyond modules capturing partial features.

To address the deficiency of human anatomical awareness in attention-based partial ReID models and aiming for efficient utilization of human body information, we propose \textbf{Part Aware Transformer (PAFormer)} with a Transformer-based structure, which incorporates pose heatmap obtained from a pose estimation model. Since PAFormer provides direct supervision of pose heatmap information to the attention map of the vanilla ViT's internal cross-attention mechanism, it is more efficient compared to other pose-based ReID methods that might rely on supplementary modules.

In detail, we introduce a novel learnable parameter called `pose token' to estimate the association between patch tokens and body parts. Unlike the part tokens in existing methods, the pose token is explicitly designed to represent a specific body part. Consequently, partial features are extracted by aggregating patch tokens based on the probability estimated by pose tokens. Additionally, we introduce a learnable visibility predictor that takes the output pose tokens as its input, in contrast to existing methods that often rely on non-learnable visibility prediction strategies involving thresholding of externally or internally estimated heatmaps.

Our contributions are summarized as follows.
\begin{itemize}

\item We point out that the lack of human anatomical awareness in previous partial ReID methods hinders the achievement of the fundamental goals of the fundamental goals of partial ReID.

\item We introduce PAFormer, which can perform part-to-part comparison by precisely localize human body parts. PAFormer introduces a novel learnable parameter called `pose token' to estimate association between body parts and patch tokens. Based on predicted association probability, PAFormer captures partial features and performs part-to-part comparison.

\item We design a visibility predictor that takes pose tokens, designed to represent each body part, as input. This predictor estimates the degree of occlusion for specific body parts. By utilizing the estimated visibility scores, we compute feature distances between samples to effectively tackle occlusion challenges.

\item PAFormer achieves state-of-the-art performance on various ReID benchmark datasets including Market-1501~\cite{Market}, DukeMTMC-ReID~\cite{Duke}, and Occluded-Duke~\cite{OCCduke}. PAFormer outperforms existing methods and demonstrates superior performance in ReID tasks.

\end{itemize}

\section{Related works}
\label{sec:sample:appendix}
\subsection{CNN based ReID methods}
Prior to the emergence of Vision Transformer, almost all ReID approaches were based on CNN. BoT~\cite{BoT} has proposed to use a CNN backbone to extract global features and to apply cross-entropy loss and triplet loss as supervision. Various subsequent researches built upon this ReID loss have also been proposed. However, because of the challenges faced by those relying soley on global features, such as occlusion, part-based approaches have emerged.

In PCB~\cite{PCB}, stripe-based methods are proposed to measure the distances between samples within the same region. Zhao \textit{et al.}\cite{Deeply} introduces CNN-based attention networks, allowing the model to autonomously find important regions in the image for ReID. ABDNet~\cite{ABDNet} builds upon this attention network concept and applies orthogonality regularization to each channel, enabling the extraction of information from various body parts. SPReID~\cite{SPReid} employs a human semantic parsing model to aggregate features only from the regions corresponding to each body part. AACN~\cite{AACN} and BPBReID~\cite{BPBReid} directly supervise formation of attention maps by utilizing pose estimation models. These approaches suggeset explicit answers on how attention maps should be formed.

\subsection{Transformer based ReID methods}
After the introduction of TransReID~\cite{TransReID}, which utilizes a jigsaw patch module to capture local features, many transformer-based ReID methods have emerged. LA Transformer~\cite{LATransformer} implements stripe-based methods on the ViT architecture. PAT~\cite{PAT} adopts an encoder-decoder structure in Transformer and uses cosine dissimilarity loss to encourage each part token to gather information from different body parts. AAFormer~\cite{AAformer} enhances distinctiveness of each token by physically limiting the number of patch tokens that can interact with each part token using Optimal Transport algorithm~\cite{optimaltransport}. NFormer~\cite{Nformer} effectively removes noise by performing self-attention only between reciprocal neighbor patch tokens. HAT~\cite{HAT} utilizes multi-scale information from ResNet~\cite{ResNet} backbone to aggregate hierarchical information. Rest-ReID~\cite{restreid} combines attention-guided graph convolutional networks with Transformer to find correlations between body parts. PFD~\cite{PFD} leverages an embedding of pose heatmaps as additional information to Transformer-based models.

\section{Problem Setting} \label{3.1}
Let us assume that $x_i$ represents the $i$-th image, and $y_i$ represents an ID label of that image. For training set $T=\{x_i, y_i\}$, a common partial ReID model is trained with ReID loss, which is a combination of cross-entropy and triplet loss. These terms can be seen as loss to solve a classification problem of determining whether $x_i$ belongs to $y_i$. With this loss, the most straightforward approach during model training would involve reducing the loss by focusing on the information from the most discriminative regions associated with each ID, which can potentially cause the model to be overfitted to the classification problem on training set $T$. Therefore, prior approaches lack the assurance that a single part prototype captures information from consistent body region across all samples and may not operate as explicitly `part-based' as they intended; instead, they could operate more as `attribute-based' (e.g., whether the image contains the color yellow).

\begin{figure}[t!]
    \begin{center}
    \includegraphics[width=\columnwidth]{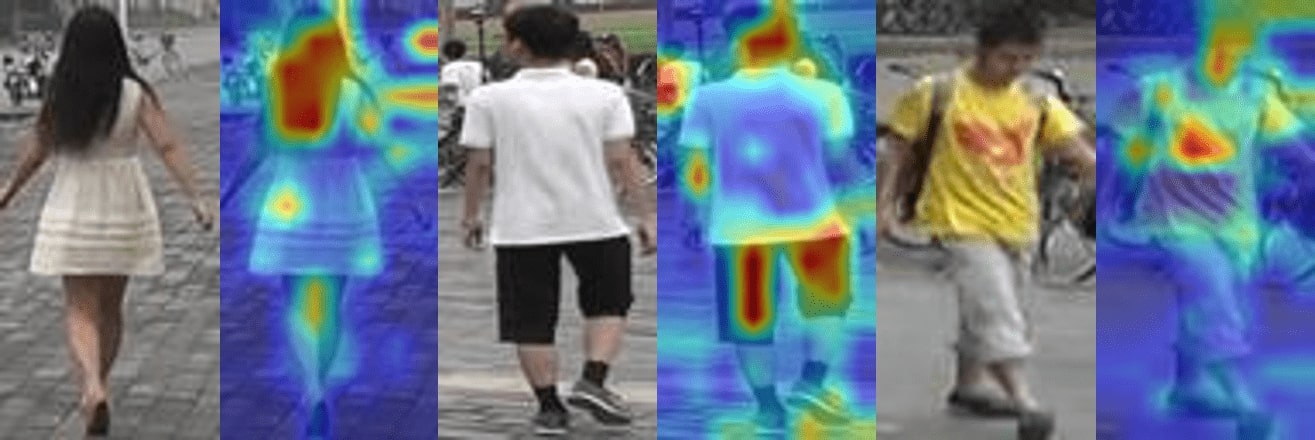}
    \end{center}
    \caption{Attention maps of vanilla ViT-based ReID model: We can observe that the first sample focuses on head, the second sample on lower body region, and the last sample on upper body region. This demonstrates that a parital ReID model trained solely on ReID loss is incapable of performing part-to-part comparisons effectively in practice.
}
    \label{Fig4} 
\end{figure}

To support our aforementioned claim, we visualize attention maps of the vanilla ViT which can be considered as a partial ReID model with a single part prototype, using attention rollout~\cite{rollout}. As shown in Figure~\ref{Fig4}, we observe that the model focuses on distinct body parts across different samples. However, since the fundamental goal of partial ReID is to perform precise comparison between same body parts, we expect the highly-focused area in attention maps to be formed on the same body parts across the samples. Highlighting the limitations of current approaches in the context of part-to-part matching, we propose our PAFormer in the following section, which enables each part prototypes to be aware of distinct human body parts.

\section{PAFormer}
\label{3.2}
\begin{figure*}[t!]
    \begin{center}
    \includegraphics[width=15cm]{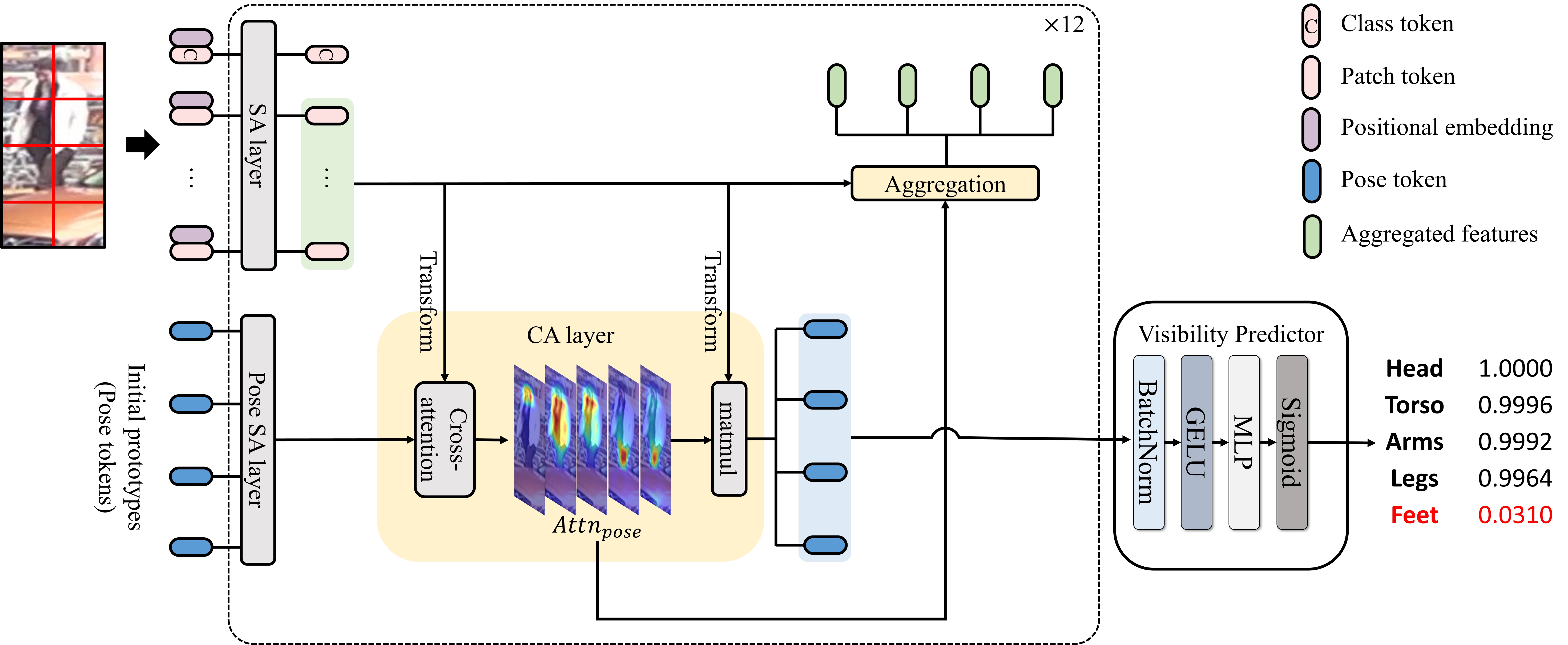}
    \end{center}
    \caption{Pipeline of PAFormer. We adopt pose tokens to estimate association between body parts and patch tokens. Partial features are generated by aggregating patch tokens during self-attention process are aggregated based on the predicted probabilities. Additionally, the output pose tokens pass through a visibility predictor to infer visibility scores.}
    \label{Fig3} 
\end{figure*}
The overall pipeline of our PAFormer is depicted in Figure~\ref{Fig3}. From section ~\ref{posetoken} to section~\ref{partialfeatures}, we provide an explanation of each component of PAFormer and elucidate the principles of the learning process. Section~\ref{3.3} describes the visibility predictor proposed to tackle occlusion issues, while section~\ref{teacherforcing} introduces modification of ReID loss based on psuedo ground truth visibility scores. Then section~\ref{3.4} covers the total loss function and equation to measure feature distance during inference. Finally, we analyze PAFormer's efficacy in section~\ref{complexity}.

\subsection{Pose tokens}\label{posetoken}
Our model employs a new learnable prototype named `pose token'. The key advantage of using a pose token is to learn to extract features of the same body part across diverse samples. This allows us to shift the main focus from aligning IDs in the training set to estimating correlations between patch tokens and body parts, which significantly aids in facilitating part-to-part comparison. Furthermore, the use of a clear guidance about human body part ensures that each pose token is trained to specifically represent a distinct body part. This attribute also enables the pose token to function as a valuable indicator for detecting occlusions in specific body parts. Detailed explanations regarding the training and application of the pose token will be provided in following subsections.

\begin{figure}[t!]
    \begin{center}
    \includegraphics[width=\columnwidth]{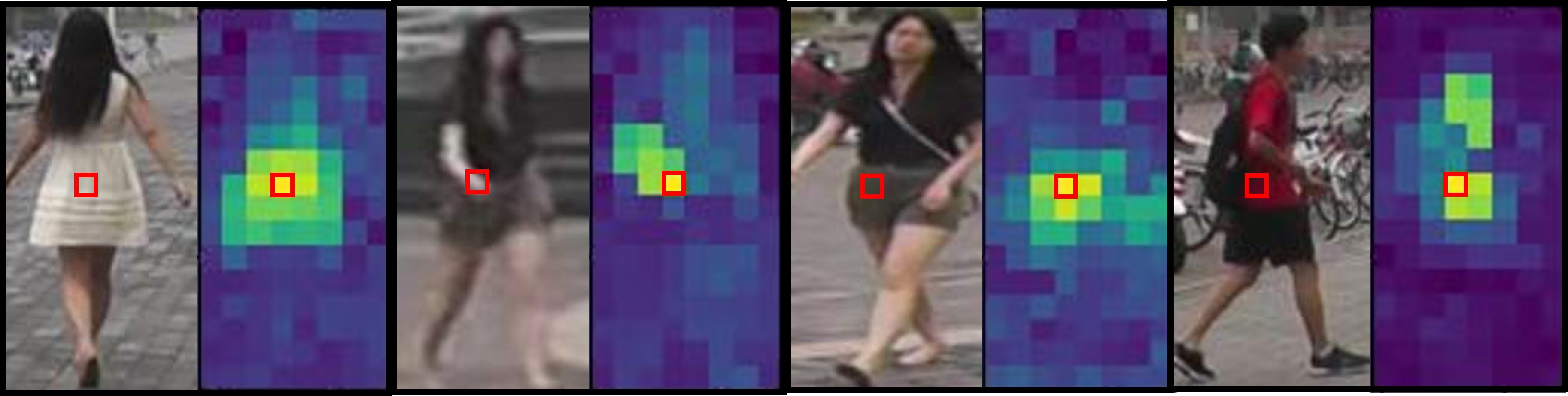}
    \end{center}
    \caption{The visualization depicts similarity between a query token (highlighted with a red border) and other patch tokens when Vanilla ViT is trained with ReID loss. Despite the consistent location of the query token, it's noticeable that patch token with high similarity varies based on the semantic of query token. For instance, in the first sample, a high similarity can be observed with tokens corresponding to the lower body, while in the second sample, tokens associated with the arms show higher similarity.}
    \label{Fig_sim} 
\end{figure}

\subsection{Feature Refinement}

We expect the pose token to demonstrate a high degree of similarity with their associated patch tokens only. We aim to leverage this similarity in the forthcoming cross-attention and aggregation layers, as detailed in subsequent subsections. To achieve this, a prerequisite step is required, which enhances the similarity among patch tokens that share the same anatomical semantics. To accomplish this, we preprocess the patch tokens using self-attention (SA), a mechanism that has already been demonstrated in various studies for effectively performing this task. In Figure~\ref{Fig_sim}, it is evident that patch tokens that share anatomical semantics converge in the vanilla ViT-based ReID model.

Additionally, by utilizing the self-attention mechanism, we can harness the robust global feature extraction capabilities of ViT. Similar to the vanilla ViT, we employ a class token to extract global features and apply ReID loss on it:
\begin{equation}
    L_{ReID}^{CLS} = L_{ID}(CLS)+L_{tri}(CLS)
\end{equation}
where $L_{ID}$ and $L_{tri}$ are cross-entropy loss and triplet loss, respectively. Also, we incorporate SA layer among pose tokens, with the objective of empowering PAFormer to learn associations among prototypes, akin to DETR~\cite{DETR} or PAT~\cite{PAT}.

\subsection{Part Awareness}
In order to imbue PAFormer with part awareness, during training, we create a ground truth heatmaps generated by post-processing initial keypoint heatmaps obtained from a pose estimation model, PifPaf~\cite{PifPaf}. The creation process of this ground truth heatmaps is detailed in section~\ref{imple}. The ground truth heatmaps directly guide how the attention maps between the pose token and patch token in the cross-attention (CA) layer should be formed. This guidance ensures that the pose token accurately perform localization of body parts. The loss function required for this process is as follows:
\begin{equation}
    L_{pose} = {1 \over P}\sum_{P}MSE(\overline{Attn}_{pose},GT)
\end{equation}
where $P$ is the number of pose tokens, $MSE$ denotes mean square error, and $GT$ represents the ground truth heatmap. $\overline{Attn}_{pose}$ is the averaged value of all $Attn_{pose}$ (12 layers$\times$12 heads) present in CA layers of PAFormer. In contrast to existing methods, PAFormer allows prototypes to concentrate on identifying patch tokens corresponding to specific body parts. 

While some methods which utilize a pose estimation model have already been proposed, they often use a separate body part localization module within their models or continue to rely on an external pose estimation model during inference. However, PAFormer takes a different approach by learning how attention maps should be formed during the cross-attention process. As a result, it does not require pose heatmaps during inference.

The underlying principle of learning part awareness through a simple mean square error is as follows:
\begin{multline}
Attn_{pose} = (X_{pose}\cdot W_{Q})\cdot(X_{patch}\cdot W_{K})^T\\
= X_{pose} \cdot (W_{Q}\cdot W_{K}^T) \cdot X_{patch}^T
= X_{pose} \cdot W \cdot X_{patch}^T
\label{exp}
\end{multline}
where $W_{Q}$ and $W_{K}$ refer to fully connected layers performing query and key transforms, respectively. $X_{pose}$ and $X_{patch}$ represents the pose tokens and the patch tokens, respectively. Expanding upon the expression, the product of $W_Q$ and $W_K$ can be expressed as a single weight matrix, denoted as $W$. The application of MSE loss offers supervision to PAFormer, allowing $W$ to accentuate channels associated with specific body parts. Consequently, the dot product of paired patch tokens and pose tokens increases, causing the attention score to selectively rise only for patches belonging to the specific body part, as intended.

\subsection{Partial Features} \label{partialfeatures}
Once the association of each pose token with specific patch tokens is estimated in the cross-attention layer, we leverage these probabilities to aggregate the refined values of the patch tokens, thereby extracting partial features. This stage is denoted as an aggregation layer. Aggregated partial features $z_{p}$ can be expressed as:
\begin{equation}
    z_{p} = \overline{Attn}_{pose}^p \cdot X_{patch}
\end{equation}
where $\overline{Attn}_{pose}^p$ is $p$-th pose token's averaged attention map across all heads from a CA layer. In typical ViT-based methods, a value transformation by a fully-connected network are applied to patch tokens when creating partial features. However, in the aggregation layer, no separate transformation is applied because the information from patch tokens should be directly accepted based on the association probabilities. Similar to the conventional Transformer architecture, as each layer is traversed, the partial features $z_{p}$ are also updated by layer normalization, feed-forward networks, and residual connections.

Relying solely on the pose heatmap for model guidance can be vulnerable if the heatmap is inaccurately generated. To mitigate this vulnerability, we supplement the guidance by incorporating the ReID loss for partial features:
\begin{equation}
    L_{ReID}^{part} = {1 \over P}{\sum_{i=1}^P}(L_{ID}(z_{i}^{L})+L_{tri}(z_{i}^{L}))
\end{equation}
where $z_{i}^{L}$ denotes the aggregated partial features from $i$-th pose token at the last PAFormer block. We minimize attention to regions that do not correspond to body parts by $L_{pose}$. Subsequently, with the assistance of $L_{ReID}^{part}$, we perform an additional filtering, addressing areas that may have been overlooked by $L_{pose}$. However, to prevent potential issues stemming from the dominant influence of the ReID loss, we assign a substantially higher weight to $L_{pose}$ than $L_{ReID}^{part}$. This strategic weighting ensures that the ReID loss functions merely as a complementary element.

\subsection{Visibility Predictor}
\label{3.3}
While many other methods that consider occlusion rely on the confidence score of the output of a module responsible for performing body part localization to address this issue, PAFormer employs a learning-based prediction network for a more sophisticated treatment of the problem, as in Figure~\ref{Fig3}. This is attributed to the design of the model in preceding stages, which ensures that the pose token serves as a representative for various body parts. For the ground truth visibility score, we assume that a body part is invisible if the maximum value in the pose heatmap does not exceed $\theta_p$. As the pose estimation model yields varying confidence scores for different body parts, $\theta_p$ is set differently for each body part. To train the visibility predictor, we employ $L_{vis}$ which uses binary cross-entropy loss between the obtained output and the ground truth:
\begin{equation}
    L_{vis} = -v_{GT}\cdot \log{v} - (1-v_{GT})\cdot\log{(1-v)}
\end{equation}
where $v$ denotes a predicted visibility score, and $v_{GT}$ denotes ground truth visibility score from the pose estimation model.

\subsection{Teacher forcing based on visibility score} \label{teacherforcing}
It is essential to refrain from calculating ID loss or triplet loss for occluded body parts. To prevent this, we adapt the ReID loss by incorporating the ground truth visibility score as a form of teacher forcing to guarantee the exclusion of occluded parts from the learning process. Particularly, during hard triplet mining~\cite{FaceNet}, to deter the selection of occluded regions, a value of 0 is multiplied to the distance for positive cases, while an extremely large value is multiplied for negative cases.

\subsection{Objective Function and Inference} \label{3.4}
The overall loss $L$ is calculated as follow:
\begin{equation}
    L = L_{ReID}^{CLS} + L_{ReID}^{part} +  \lambda L_{pose} + L_{vis} 
\end{equation}
$\lambda$ is a weight for $L_{pose}$ and is heuristically set to 10. All modules in PAFormer are learned together.

During inference phase, PAFormer calculate distances as follows:
\begin{equation}
    d^{i,j} = d_{CLS}^{i,j} + {   \sum_{p=1}^{P} d_{p}^{i,j}v_{p}^{i}v_{p}^{j} \over   \sum_{p=1}^{P} v_{p}^{i}v_{p}^{j}       }
\label{distance}
\end{equation}
where $d^{i,j}$ is a distance between $i$-th and $j$-th samples, and $v$ denotes the predicted visibility score. Among the subscripts in $d$, `CLS' refers to the class token, and $p$ indicates the specific pose token's identifier.

\subsection{Time complexity of PAFormer} \label{complexity}
Generally, time complexity of Transformer is known to be $O(N^2d)$, where $N$ and $d$ are the number of tokens and channels, respectively. Since PAFormer performs self-attention similarly, it maintains the same $O(N^2d)$ time complexity. Additionally, the pose token self-attention process contributes $O(P^2d)$, and the cross-attention process adds $O(PNd)$, resulting in an overall time complexity of $O((N^2 + P^2 + PN)d)$. Given that $N$ is usually much larger than $P$ ($N$: three-digit, $P$: one-digit), the additional burden introduced by PAFormer is considered to be negligible.

\section{Experiments}

\subsection{Datasets}

\begin{table}[hbt!]
\centering
\begin{tabular}{cccc}
\hline
Dataset       & \#image & \#ID  & \#cam \\ \hline

Market-1501~\cite{Market}   & 32,668  & 1,501 & 6     \\
DukeMTMC-ReID~\cite{Duke} & 36,441  & 1,404 & 8     \\
Occluded-Duke~\cite{OCCduke} & 36,441  & 1,404 & 8     \\ \hline

\end{tabular}
\caption{Statistics of ReID datasets used in experiments.}
\label{Dataset statistics}
\end{table}


Three widely used ReID datasets are chosen for our experiments and their statistics are shown in Table~\ref{Dataset statistics}. Images in every dataset are resized to $256\times 128$ or $384\times 128$. Then, the training images are augmented with random horizontal flipping, cropping, padding, grayscale~\cite{Grayscale} and erasing~\cite{randomerasing}. Among these methods, we apply all except random erasing and grayscale augmentation to the pose mask. Additionally, for regions affected by random erasing in the original data, we set the mask value to 0.


\subsection{Implementation Details} 
We adopt ImageNet pre-trained ViT-B/16 as the backbone. The model is trained with SGD optimizer with a momentum of 0.9 and a weight decay of 1e-4. The learning rate is set to 0.008 with a cosine decay and the batch size is determined to be 64. We also use sliding-window ($S=12$) and side information embedding~\cite{TransReID}. Both training and testing processes are implemented with two NVIDIA GeForce RTX 4080 GPUs. We conducted training for 320 epochs and evaluated the performance every 20 epochs. Among these evaluations, we choose the one with the highest performances.  

\begin{table}[t!]
\centering
\begin{tabular}{clccccc}
\hline
\multicolumn{1}{l}{} &                      & Head & Torso & Arms & Legs & Feet \\ \cline{1-7} 
\multicolumn{2}{c}{Market-1501}             & 0.6     &  0.7     &  0.85    & 0.8     &   0.7   \\
\multicolumn{2}{c}{Duke-reID}               & 0.6  & 0.8   & 0.85 & 0.85 & 0.75 \\
\multicolumn{2}{c}{OCC-Duke}                & 0.6  & 0.8   & 0.85 & 0.85 & 0.75  \\ \hline
\end{tabular}
\caption{Values of $\theta_{p}$ for various datasets.}
\label{lambdap}
\end{table}

\subsection{Ground truth Pose heatmap ground truth Visiblity score}\label{imple}
We use PifPaf~\cite{PifPaf} to obtain pose heatmaps from the input image. However, instead of using the heatmaps predicted by PifPaf directly, we employ several modification steps to adapt them as ground truth heatmaps.

Firstly, we apply random erasing augmentation~\cite{randomerasing} to pose heatmap as well. The probabilities corresponding to the erased areas in original image are all set to 0.

Secondly, we combine heatmaps of small fragments such as keypoints and joints to form masks for various body parts. During this process, we regard heatmaps from PifPaf with a maximum confidence score below $\theta_p$ as noise and ignore them. The value of $\theta_p$ is set differently for each body part and for each dataset, as in Table.~\ref{lambdap}. 

Thirdly, we normalize the combined heatmaps. Given our objective is to avoid focusing solely on discriminative parts, we assume that the influence of all patch tokens belonging to a specific body part is uniform. Consequently, value in the ground truth mask is either 0 or ${1 \over K}$, where $K$ represents the count of non-zero values in the ground truth mask. Following this transformation, we normalize the mask by dividing it by the sum of its constituent values, ensuring a total sum of 1.

The ground truth visibility score is also determined by $\theta_p$. Pseudo labels are assigned as `visible' only for body parts with a maximum confidence score greater than or equal to $\theta_p$.

\subsection{Visualization}

\begin{figure*}[t!]
    \centering

    \includegraphics[width=\textwidth]{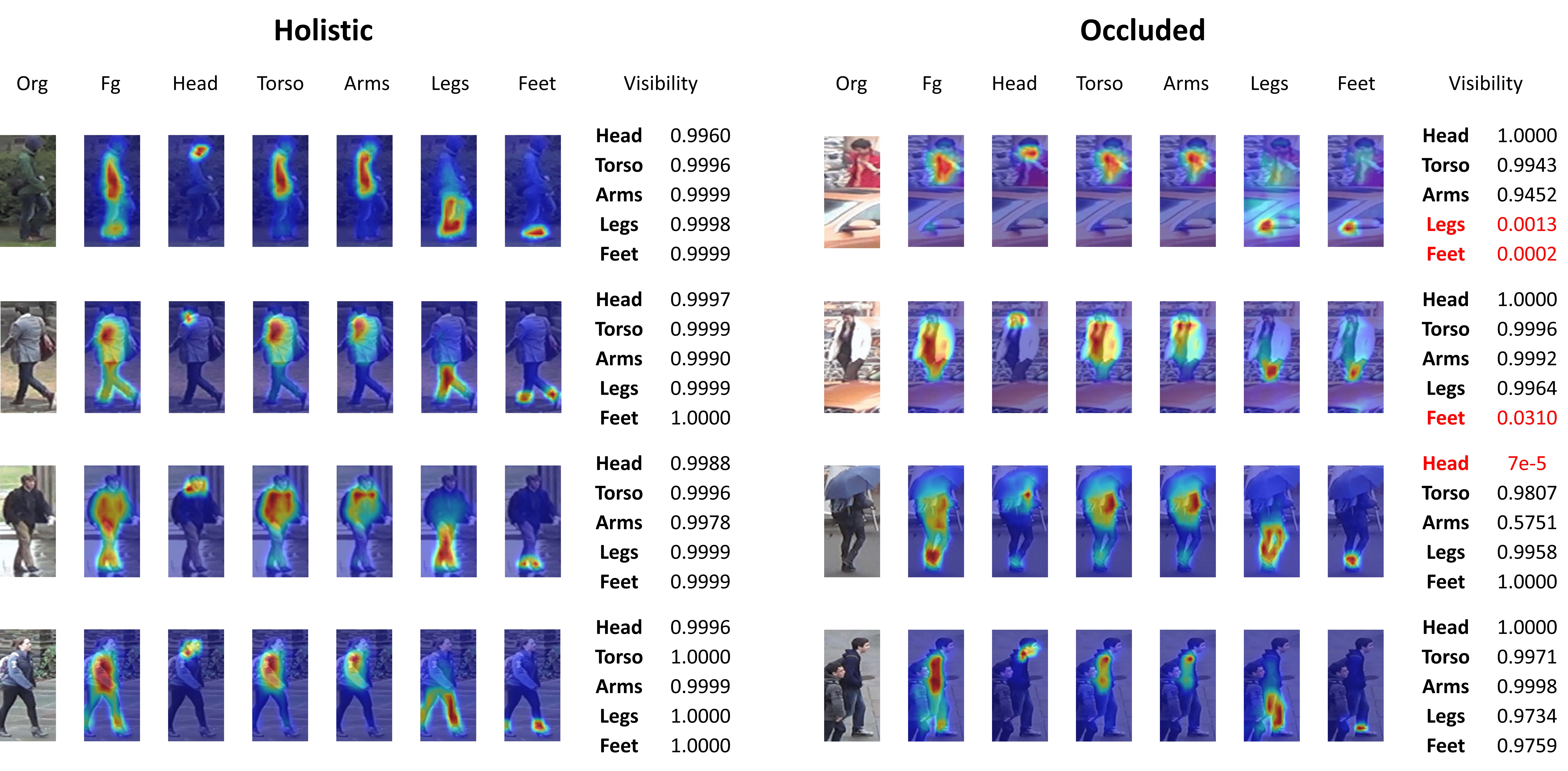}
    \caption{Attention maps and visibility scores of pose tokens. The color of the attention map goes from blue to red as attention score goes higher. The example divides a person into five body parts ($P=5$), and `Fg' refers to foreground features that combine all of the parts. Even as samples change, it's evident that each pose token continues to focus on the same body part it represents. In addition, we show a visibility score of each pose token on the right side of attention maps. It is observed that the visibility score corresponding to occluded body part is predicted very low (The scores predicted low are highlighted in red). }
    \label{Fig_attvis}

\end{figure*}
\subsubsection{Attention maps} We present attention maps of the pose tokens in Figure~\ref{Fig_attvis}, visualized by averaging the attention maps across all CA layers. The visualization highlights that the pose tokens consistently concentrate on their respective distinctive body parts across various samples. This starkly contrasts with the attention maps in Figure~\ref{Fig4}, which are solely learned through ReID loss. Figure~\ref{Fig2} emphasizes that each prototype should aggregate information from each patch token based on their semantics, regardless of ID-specific appearances. Considering that the attention map in CA layer represents a similarity matrix between pose token and patch tokens, the visualization in Figure~\ref{Fig_attvis} demonstrates the successful resolution of this issue. 

Furthermore, in instances where only a portion of a specific body part is occluded, PAFormer demonstrates the ability to selectively focus on the visible regions (Please refer to the Legs section of the second sample in the occluded part, in Figure~\ref{Fig_attvis}). When a particular body part is fully occluded, PAFormer tends to concentrate on regions where corresponding body parts might be located (See Legs and Feet section of the first sample in the occluded part, in Figure~\ref{Fig_attvis}). However, this is subsequently adjusted through visibility prediction.

\subsubsection{Visibility scores} To further validate the accuracy of the predicted visibility scores, we also provide scores measured for samples, as shown in Figure~\ref{Fig_attvis}. The scores in the figure are sigmoid-activated, clearly indicating that occluded body parts lead to notably lower visibility scores. Based on this, we measure distances between samples as decribed in \eqref{distance}, effectively mitigating the detrimental impact of occlusion.

Attempts to convert visibility scores into binary values of 0 or 1 resulted in a significant performance decrease. This highlights the advantage of depending on the model's assessment of visibility, rather than entirely disregarding ambiguously visible regions (e.g., cases with visibility scores ranging between 0.2 and 0.8).

\begin{table*}[hbt!]
\centering
\resizebox{0.9\textwidth}{!}{
\begin{tabular}{ccccccccc}
\hline
\multirow{2}{*}{Method} & \multirow{2}{*}{Ref} & \multirow{2}{*}{Size} & \multicolumn{2}{c}{Market-1501} & \multicolumn{2}{c}{Duke-reID} & \multicolumn{2}{c}{OCC-Duke} \\
                                            &       &     & mAP  & R1   & mAP  & R1   & mAP  & R1    \\ \hline
PCB+RPP~\cite{PCB}                          & ECCV18& 256 & 81.6 & 93.8 & 69.2 & 83.3 & -    & -     \\
*ISP~\cite{ISP}                              & ECCV20& 256 & 88.6 & 95.3 & 80.0 & 89.6 & 52.3 & 62.8  \\
CBDB-Net~\cite{CBDB}                        & CSVT21& 256 & 85.0 & 94.4 & 74.3 & 87.7 & 38.9 & 50.9   \\
CDNet~\cite{CDNet}                          & CVPR21& 256 & 86.0 & 95.1 & 76.8 & 88.6 & -    & -     \\
C2F~\cite{C2F}                              & CVPR21& 256 & 87.7 & 94.8 & 74.9 & 87.4 & -    & -     \\
PAT~\cite{PAT}                              & CVPR21& 256 & 88.0 & 95.4 & 78.2 & 88.8 & 53.6 & 64.5   \\
TransReID~\cite{TransReID}                  & ICCV21& 256 & 88.9 & 95.2 & 82.0 & 90.7 & 59.2 & 66.4   \\
HAT~\cite{HAT}                              & ACM21 & 256 & 89.5 & 95.6 & 81.4 & 90.4 & -    & -  \\
*PFD~\cite{PFD}                              & AAAI22 & 256 & 89.7 & 95.5 & 83.2 & \underline{91.2} & \textbf{61.8} & \textbf{69.5} \\
ResT-ReID~\cite{restreid}                   & PRL22 & 256 & 88.2 & 95.3 & 80.6 & 90.0 & 51.9 & 59.6     \\
DCAL~\cite{DCAL}                            & CVPR22& 256 & 87.5 & 94.7 & 80.1 & 89.0 & -    & -     \\
ABDNet~\cite{ABDNet}+NFormer~\cite{Nformer} & CVPR22& 256 & \textbf{93.0} & 95.7 & \textbf{85.7} & 90.6 & -    & -     \\
*BPBReID~\cite{BPBReid}                      & WACV23& 256 & 87.0 & 95.1 & 78.3 & 89.6 & 54.1 & 66.1     \\
DCFormer~\cite{DCFormer}                    & AAAI23& 256 & 90.4 & \textbf{96.0} & - & - & - & -     \\ \hline
\textbf{PAFormer}                           &-      & 256 & \underline{90.8} & \textbf{96.0} & \underline{83.3} & \textbf{92.1} & \underline{59.9} & \underline{67.0} \\ \hline
ABDNet~\cite{ABDNet}                        &ICCV19 & 384 & 88.3 & 95.6 & 78.6 & 89.0 & -    & -     \\
OH-former~\cite{Ohformer}                   &Arxiv21& 368 & 88.7 & 95.0 & 82.8 & 91.0 & -    & -     \\
AAformer~\cite{AAformer}                    &Arxiv21& 384 & 87.7 & 95.4 & 80.0 & 90.1 & 58.2 & \underline{67.0}  \\
TransReID~\cite{TransReID}                  & ICCV21& 384 & 89.5 & 95.2 & 82.6 & 90.7 & 59.4 & 66.8 \\
MGN~\cite{MGN}+AutoLoss~\cite{Autoloss}     & CVPR22& 384 & \underline{90.1} & \textbf{96.2} & -    & -    & -    & -    \\
*BPBReID~\cite{BPBReid}+HRNet~\cite{HRNet}   & WACV23& 384 & 89.4 & 95.4 & \textbf{84.2} & \underline{92.4} & \textbf{62.5} & \textbf{75.1}   \\ \hline
\textbf{PAFormer}                           &-      & 384 & \textbf{90.9}    & \underline{96.1}    & \textbf{84.2} & \textbf{92.5} & \underline{60.4}    & 66.4     \\ \hline
\end{tabular}
}
\caption{Performance comparison with state-of-the-art(SOTA) methods on ReID benchmarks. * denotes pose-estimation based methods. The highest and second-highest scores for each dataset are denoted by bold letters and underlines, respectively. Since TransReID~\cite{TransReID} does not provide its performances on Occluded-Duke at the size of $384\times128$, we reproduced the results. Performance results are provided for both $256\times128$ and 368 or $384\times128$ resized inputs. Our PAFormer either outperforms existing SOTA models or shows performance on par with them.}
\label{Performances}
\end{table*}

\subsection{Comparison with existing methods}
Table~\ref{Performances} shows the performances of our PAFormer and other previous ReID methods. We adopt mean Average Precision (mAP) and Rank-1 score (R1) for evaluations. PAFormer demonstrates its strong capabilities by achieving state-of-the-art performance on various datasets, outperforming competitors and securing at least the second-place position. The performances using the holistic datasets and the occluded dataset have been achieved with $P=5$ and $P=6$, respectively. 
\\

\noindent \textbf{Market-1501.} For image size of $256\times128$, PAFormer achieves 90.8/96.0 for mAP and Rank-1, respectively. In terms of the Rank-1 score, PAFormer exhibits the highest score among the entire comparison group, and the mAP also surpasses 90, showing the second-highest performance following Nformer~\cite{Nformer}. Since NFormer uses ABDNet~\cite{ABDNet}, which has already demonstrated excellent performance in the ReID field, as its backbone, PAFormer appears to lag slightly behind in terms of mAP.

PAFormer exhibits strong performance even for 384$\times$128 images, achieving 90.9 in mAP and 96.1 in Rank-1 score. It records the highest mAP and the second-highest Rank-1 score among the comparison groups, demonstrating excellent performance.
\\

\noindent \textbf{DukeMTMC-reID.} Our method demonstrates a performance of 83.3 in mAP and 92.1 in Rank-1 on the DukeMTMC-ReID dataset with a resolution of $256\times128$. These scores correspond to the 2nd and 1st places, respectively, among the entire comparison group. While NFormer~\cite{Nformer} demonstrates impressive performance in terms of mAP on holistic datasets, PAFormer exhibits superior results in terms of Rank-1 scores. 

When applied to images resized to $384 \times 128$, PAFormer achieves a performance of 84.2 in mAP and 92.5 in Rank-1 score, which are both the highest scores among the comparisons. The high performance on both Market-1501 and DukeMTMC-reID datasets demonstrates the ReID ability of PAFormer on holistic datasets.
\\

\noindent \textbf{Occluded-Duke.} To evaluate performance on occluded cases, we experiment on Occluded-Duke dataset. For images of size 256$\times$128, PAFormer achieves mAP of 59.9 and Rank-1 of 67.0. PAFormer achieves the second highest performance on both mAP and Rank-1 score. Considering that PFD~\cite{PFD} uses the pose estimation model during inference, PAFormer's performance is competitive enough. While showing a slightly weaker performance on 384$\times$128 images, PAFormer still achieves the second-highest mAP among the comparison groups. 
The decline in performance at image size of 384$\times$128 can be attributed to the growing inaccuracy of the pose estimation model's heatmaps in occluded dataset as the image size increases.

\subsection{Ablation Study}
\label{Ablation}
\begin{table}[t!]
\centering
\begin{tabular}{c|ll}
\hline
$P$ & \multicolumn{2}{c}{Grouping strategy}                                    \\ \hline
3 & \multicolumn{2}{l}{\{Head, Upper, Lower\}}                               \\
4 & \multicolumn{2}{l}{\{Head, Upper, Legs, Feet\}}                          \\
5 & \multicolumn{2}{l}{\{Head, Torso, Arms, Legs, Feet\}}                    \\
6 & \multicolumn{2}{l}{\{Heat, Upper torso, Lower torso, Arms, Legs, Feet\}} \\ \hline
\end{tabular}
\caption{Grouping strategy of body parts for different $P$.}
\label{AblationDiv}
\end{table}

\begin{figure}[t!]
    \begin{center}
    \includegraphics[width=\columnwidth]{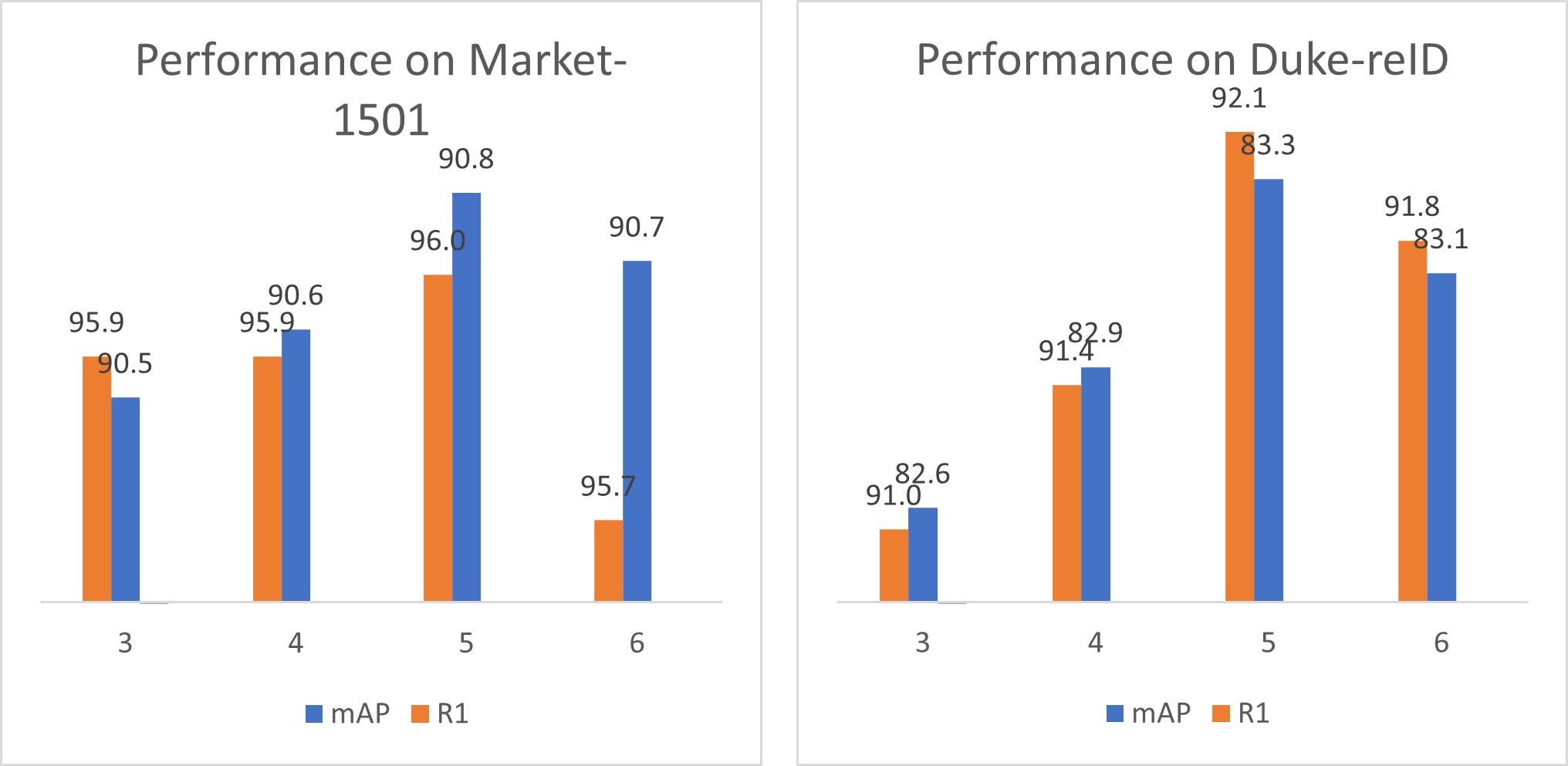}
    \end{center}
    \caption{Performance comparison for different number of pose tokens, $P$.  }
    \label{perfortable} 
\end{figure}
\subsubsection{The number of pose tokens, $P$} In order to analyze the impact of body part division methods on performance, we conduct an ablation study for different values of $P$. Table~\ref{AblationDiv} illustrates body part dividing strategies according to various $P$, while Figure~\ref{perfortable} displays the corresponding performance outcomes. We confirmed that the optimal value of $P$ for the holistic datasets is 5.

When the value of $P$ is too small, it becomes hard to handle occlusion problem. For instance, when the value of $P$ is 4, since the torso and arms are not be separated, the features of the torso might be influenced even in cases where the arm region is occluded. 
Moreover, there may be disadvantages in situations where there is a significant difference in features between the arms and torso, for example, when the individual is wearing short sleeves or sleeveless clothing.

On the contrary, when $P$ is excessively large, there is a risk of overfitting. Additionally, the influence of a particular body part may be duplicated during feature distance calculation, leading to an unintended amplification of the impact of that specific body parts.

\begin{table}[t!]
\centering
\begin{tabular}{ccc}
\hline
visibility score & mAP  & R1   \\ \hline
w             & \textbf{59.9} & \textbf{67.0} \\
w/o               & 57.6 & 64.5 \\
round            & 59.0 & 66.1 \\ \hline
\end{tabular}
\caption{Ablation of the visibility score.}
\label{visablation}
\end{table}
\subsubsection{Validity of the visibility score} In order to assess the effectiveness of the visibility prediction, we conducted an ablation study on visibility scores, in Table~\ref{visablation}. We experiment on the three scenarios for the visibility scores predicted by the model: 1) using the visibility score as-is, 2) excluding the visibility score, and 3) rounding the visibility score. All experiments are conducted on the Occluded-Duke with $P=6$. It is confirmed that using the visibility score from PAFormer as-is yielded the best results.

\begin{figure}[t!]
    \begin{center}
    \includegraphics[width=\linewidth]{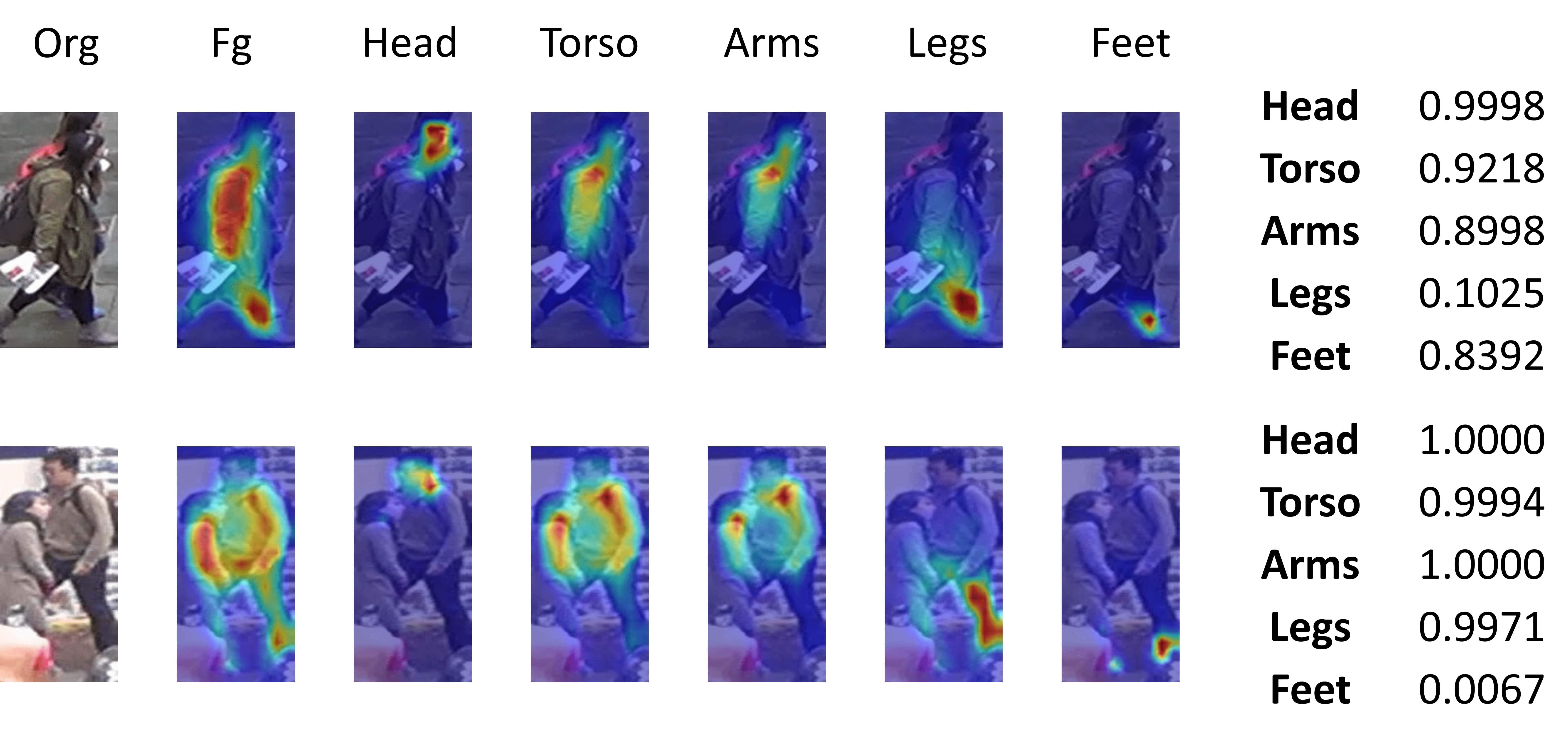}
    \end{center}
    \caption{Failure cases: In cases where the visibility score or attention region is inaccurately formed. For instance, in the first sample, even though the legs are visible, the visibility score has been measured to be significantly low. In the second sample, the model is focusing on wrong areas. 
 }
    \label{Fig_fail} 
\end{figure}

\subsection{Limitation} PAFormer's training strategy hinges on the output pose heatmap from a pose estimation model, revealing vulnerabilities in scenarios with two or more individuals, as depicted in Table~\ref{Fig_fail}. This challenge is not so much a flaw in our methodology but rather stems from inaccuracies in generating ground truth masks. By using models adept at multi-person pose estimation and leveraging precisely annotated data, PAFormer has the potential to achieve enhanced performance in such situations.

\section{Conclusion}
This paper introduces an impactful model for partial ReID, named PAFormer. We tackle the shortcomings observed in existing partial ReID models by proposing PAFormer, which is aware of anatomical aspect of body parts. Our approach involves the use of learnable pose tokens to estimate the correlation between patch tokens and different body parts. Subsequently, we aggregate information from patch tokens based on the attention maps generated by the pose tokens. Furthermore, we introduce a visibility predictor to effectively handle occlusion issues. PAFormer demonstrates state-of-the-art performance on well-known ReID datasets.

\bibliographystyle{elsarticle-num-names} 
\bibliography{cas-refs}





\end{document}